\documentclass[letterpaper, 10 pt, conference]{ieeeconf}  % Comment this line out if you need a4paper
\usepackage{graphicx}

\IEEEoverridecommandlockouts                              % This command is only needed if 
\usepackage{amsmath} % assumes amsmath package installed
\usepackage{amsfonts}  % assumes amsmath package installed
\usepackage{array} % temp
\usepackage{multirow} % temp
\usepackage{caption}
\usepackage{booktabs}
\usepackage{commath}
\usepackage{mathtools}
\usepackage{hyperref}
\usepackage{gensymb}

\title{\LARGE \bf
Data-driven Loop Closure Detection in Bathymetric Point Clouds for Underwater SLAM
}

\author{Jiarui Tan$^{1}$, Ignacio Torroba$^{1}$, Yiping Xie$^{1}$, and John Folkesson$^{1}$% <-this % stops a space
\thanks{The authors are with the \href{https://www.kth.se/is/rpl/division-of-robotics-perception-and-learning-1.779439}{Division of Robotics, Perception and Learning} at KTH Royal Institute of Technology, SE-100 44 Stockholm, Sweden.
{\tt\small \{jiaruit, torroba, yipingx, johnf\}@kth.se}}}%

\begin{document}

\maketitle
\thispagestyle{empty}
\pagestyle{empty}

%%%%%%%%%%%%%%%%%%%%%%%%%%%%%%%%%%%%%%%%%%%%%%%%%%%%%%%%%%%%%%%%%%%%%%%%%%%%%%%%
\begin{abstract}
Simultaneous localization and mapping (SLAM) frameworks for autonomous navigation rely on  robust data association to identify loop closures for back-end trajectory optimization. In the case of autonomous underwater vehicles (AUVs) equipped with multibeam echosounders (MBES), data association is particularly challenging due to the scarcity of identifiable landmarks in the seabed, the large drift in dead-reckoning navigation estimates to which AUVs are prone and the low resolution characteristic of MBES data. Deep learning solutions to loop closure detection have shown excellent performance on data from more structured environments. However, their transfer to the seabed domain is not immediate and efforts to port them are hindered by the lack of bathymetric datasets. Thus, in this paper we propose a neural network architecture aimed to showcase the potential of adapting such techniques to correspondence matching in bathymetric data. We train our framework on real bathymetry from an AUV mission and evaluate its performance on the tasks of loop closure detection and coarse point cloud alignment. Finally, we show its potential against a more traditional method and release both its implementation and the dataset used.
\end{abstract}

%%%%%%%%%%%%%%%%%%%%%%%%%%%%%%%%%%%%%%%%%%%%%%%%%%%%%%%%%%%%%%%%%%%%%%%%%%%%%%%%
\section{INTRODUCTION}
There is an increasing demand from both the scientific community and industry to create high resolution terrain models of the ocean floor. Tasks such as laying of pipelines and cables or deep sea oceanographic and environmental research necessitate accurate seabed maps for safe execution and high-fidelity geo-referencing of the data collected \cite{graham2022rapid}. Autonomous underwater vehicles (AUVs) equipped with multibeam echosounders sonars (MBES) are becoming essential tools for these missions, thanks to the high-resolution 3D reconstructions of the underwater environment that they can provide, regardless of the water conditions.  This is due to their capacity to get closer to the seafloor and to reach inaccessible  areas compared to surface vessels. However, their ability to travel long distances underwater is hindered by the lack of a global positioning system (such as GPS) underwater that could bound the navigation drift to which AUVs are particularly sensitive. Although similar systems exist for underwater vehicles, such as long baseline (LBL) or ultra-short baseline (USBL) \cite{rigby2006towards}, they are limited in their operational capabilities by the propagation of sound in water and the need to deploy the transponders close enough to the the survey site.

In this context, simultaneous localization and mapping (SLAM) techniques can provide an alternative method for the vehicle to autonomously correct its trajectory estimate based only on its own proprioceptive sensors and the measurements collected from the environment \cite{leonard2016autonomous}. Current state-of-the-art bathymetric SLAM frameworks fuse dead reckoning (DR) estimates and trajectory corrections from the registration of overlapping submaps in a factor graph optimization \cite{torroba2019towards}. Although well-studied, these techniques rely critically on the successful detection of the overlap between submaps upon loop closure (LC), which remains an open problem for bathymetric data in large, unstructured environments \cite{hammond2015automated}. This is mainly due to the scarcity of distinguishable landmarks on the seabed, together with the low resolution and the noise levels characteristic of MBES data. 

Thus, this work focuses on the problem of autonomous data association in bathymetric point clouds.  Given the complexity of deriving principled solutions to this problem in unstructured environments, we present a data-driven approach built upon the latest advancements in keypoint selection and descriptor learning achieved by neural network architectures \cite{yew20183dfeat, lu2019deepvcp}. We train and evaluate our method on multibeam echosounder data from an AUV mission under the Thwaites glacier in Antarctica.   Finally, although our results are limited by the amount of data available, we show the potential for our approach to be readily applied to loop closure detection and coarse alignment in submap-based bathymetric SLAM frameworks.

\section{RELATED WORK}
Autonomous place recognition and point cloud alignment are two key components of SLAM solutions as they provide loop closure constraints for bounding the vehicle's DR drift. Both processes rely on keypoint selection and descriptor embedding to detect and describe salient points that can be easily disambiguated. These descriptors are then used in correspondence matching for place recognition and coarse alignment of the point clouds.

Regarding the selection of keypoints and the construction of their descriptors, the existing approaches can be divided into handcrafted and learning-based. 
As examples of the former, in \cite{hammond2015automated} the bathymetric point clouds are converted to "pseudo-images" to extract SIFT descriptors \cite{lowe1999object} for the matching. Working directly on the point clouds, \cite{suresh2020active} extracts keypoints  with a Harris 3D detector \cite{sipiran2011harris}. These are then encoded in SHOT descriptors \cite{salti2014shot}, which are clustered to create bathymetric submap dictionaries for loop closure detection. This method will be applied later in this paper as a baseline. 
Data-driven methods go beyond manual design of detectors and descriptors and instead aim to learn them directly from the data. This was made possible by the seminal PointNet from \cite{qi2017pointnet}, which allowed neural networks to work directly with geometric point clouds. This architecture and its extension to hierarchical learning, PointNet++ \cite{qi2017pointnet++}, enabled further research on learning local descriptors directly from point clouds \cite{deng2018ppfnet, yew20183dfeat, lu2019deepvcp}. \cite{deng2018ppfnet} proposed a novel N-tuple loss based on the contrastive loss from \cite{1640964}, combining point pair features and global context to learn a globally-aware feature descriptor. \cite{lu2019deepvcp} presented a successful end-to-end point cloud registration framework, although constrained by the need of accurate ground truth poses in the training stage. To overcome this, in \cite{yew20183dfeat}, a weakly-supervised framework is designed to learn salient local features from 3D point clouds without precise knowledge of the ground truth point-to-point correspondence matching. 
Moving beyond neural networks, in \cite{hitchcox2020point} Gaussian processes (GP) \cite{rasmussen2003gaussian} are used to find salient keypoints by comparing the smooth GP posterior mean trained on a patch of bathymetry against the raw data.

Finding correspondences across point clouds in the feature space then allows to perform place recognition. Random sample consensus (RANSAC) \cite{fischler1981random} has been widely employed for this due to its simplicity and robustness against outliers \cite{raguram2008comparative}. In \cite{dube2017segmatch}, a random forest is trained to learn the most representative metric to perform matching in their highly-complex feature space. In \cite{hitchcox2020point}, a target and a source graph are constructed from the keypoints and the matching is carried out as a weighted network alignment problem.
The correspondences found can then be used to define a coarse alignment between point clouds that will be used as the initial pose in the fine registration step. The iterative closest point (ICP) \cite{besl1992method} and its variants have been widely used in SLAM frameworks \cite{torroba2019towards, suresh2020active}. Data driven approaches exist that seek to learn the full registration process from a pair of point clouds \cite{lu2019deepvcp, aoki2019pointnetlk}. However, following \cite{torroba2018comparison} in this work we employ the generalized ICP (GICP) \cite{segal2009generalized} method for fine registration of the bathymetric point clouds in the experimental section.

From the works presented above, only \cite{hammond2015automated, suresh2020active} and \cite{hitchcox2020point} target bathymetric data, which already indicates the complexity of the problem. However, in \cite{suresh2020active} the data is collected in an structured environment and in \cite{hitchcox2020point} a high-resolution subsea optical laser scanner is used. In our work, we have focused on the problem of keypoint selection and feature description MBES sonar data, the most common sensor used in large-scale, deep sea bathymetric surveying nowadays. 

\section{OUR METHOD}
Given a set of $N$ $V$-dimensional points, $P_i\in \mathbb{R}^{N\times V}$, the goal is to characterize their $W$-dimensional descriptors $\xi_i \in \mathbb{R}^{N\times W}$ such that the most representative geometric features common across overlapping point clouds result in similar descriptors that can be then matched for loop closure detection. More specifically, we aim to create point descriptors that encapsulate local information and that are robust against noise. Furthermore, in order to quantify their saliency, we seek to define a set of scalars $w_i \in \mathbb{R}^{N}$ that weight the descriptors' prominence within their environment. Specifically, we make use of the Siamese architecture \cite{1467314} to learn these descriptors and their weights from $P_i$.

\begin{figure}[t]
  \centering
  \includegraphics[width=0.25\textwidth,angle =90]{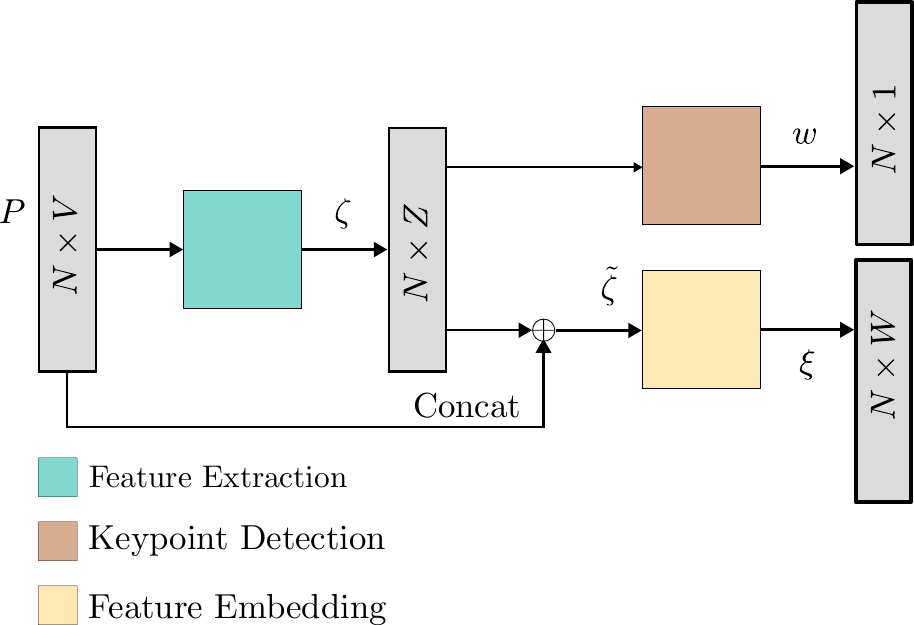}
  \caption{Our inference network architecture. Point clouds are fed into the network and weights $w$ and descriptors $\xi$ are output. $w$ indicate the saliency of the features and $\xi$ are used for down-stream data association.}
  \label{fig:nn-pipeline-inference}
\end{figure}

\subsection{Metric learning}
Formally, we aim to regress a mapping of the form $f: P_i \mapsto \xi_i, w_i$ parameterized by a set of variables $\theta$ so that the common features between $\xi_i$ and $\xi_j$, within an overlapping area from a second set of points $P_j$, are close to each other according to some metric. Additionally, we intend the value of $w_i$ to directly correlate with the saliency of the descriptors within $\xi_i$. Since the mapping $f$ is intractable, learning such function can be formulated as a metric learning problem \cite{weinberger2009distance}.
The triplet loss approach \cite{schroff2015facenet} to metric learning aims to construct a latent space where similar descriptors are brought closer and different ones farther, thus defining a distance metric that encodes the similarity of the geometric features within the space. This is achieved by working on sets of inputs or $triplets$, which consist of an anchor input ($a$) randomly sampled, a positive one ($p$) known to be similar to the anchor and a negative one ($n$), with nothing in common with the anchor.
Eq. \ref{eq:tri_loss_ind} shows the standard triplet loss for a triplet $\{a,p,n\}_t$.

\begin{equation}
    \mathcal{L}_t = \big[\norm[0]{ {\xi}_a-{\xi}_p}_2 - \norm[0]{ {\xi}_a-{\xi}_n}_2 +\gamma \big]_{+},
    \label{eq:tri_loss_ind}
\end{equation}
where $[z]_+=\max(z,0)$ denotes the hinge loss, $\norm[0]{\cdot}_2$ is the L2-norm and $\gamma$ models the margin enforced between the distances of positive and negative pairs. In our case, the triplets comprise sets of points $P_i$ and the similarity between sets is measured in terms of physical overlap.

\subsection{Proposed architecture}
Learning $f$ can be achieved by maximizing the likelihood of its parameters $\theta$ for a given dataset of triplets. To this end, we divide the process into three learning modules denoted: i) feature extraction $\phi$, ii) keypoint detection $\psi$ and iii) feature embedding $\chi$, as shown in Fig \ref{fig:nn-pipeline-inference}.

As a first step, for a given input point set $P_i = \{ p_{k}\}_{k=1}^N $, the deep feature extraction network learns a $Z$-dimensional feature $\zeta_k \in \mathbb{R}^Z$ for each of the input points. Using the resulting $\zeta_k$ the keypoint detector module $\psi$ further maps each descriptor to its scalar weight $w_k \in \mathbb{R}$, which models the salience of the original point.
In parallel, each $\zeta_k$ descriptor is then concatenated with the relative position of its corresponding point (with respect to the centroid of the point clouds), that is, $\Tilde{\zeta}_k = \zeta_k \oplus p_{k}$, to include geometric information. Using $\Tilde{\zeta}_k$, the feature embedding network $\chi$ seeks to embed each $\Tilde{\zeta}_k$ into a more detailed $W$-dimensional feature descriptor $\xi_k \in \mathbb{R}^{W}$ that includes information from the neighbouring points.

The original $\theta$ is now composed of three sets of learnable parameters $\theta = \{ \theta_\phi$, $\theta_\psi , \theta_\chi \}$ parameterizing the three neural networks, $\phi$, $\psi$ and $\chi$ respectively. The original function $f: \mathbb{R}^{N\times V} \to    \mathbb{R}^{N \times W} \times \mathbb{R}^N $ can now be described in more detail as follows: \begin{equation}
    f (P_i,\theta_\phi, \theta_\psi, \theta_\chi) =(\pi_1(P_i,\theta_\phi,\theta_\chi),\pi_2(P_i,\theta_\phi,\theta_\psi))
\end{equation}
\begin{equation}
    \pi_1(P_i, \theta_\phi,\theta_\chi)=\chi(\phi(P_i), P_i) : \mathbb{R}^{N\times V} \to \mathbb{R}^{N\times W},
\end{equation}
\begin{equation}
    \pi_2(P_i, \theta_\phi,\theta_\psi)=\psi(\phi(P_i)) : \mathbb{R}^{N\times V} \to \mathbb{R}^N.
\end{equation}

The details of each module are described below. 

\subsubsection{Deep feature extraction}
The hierarchical propagation strategy adopted in \cite{qi2017pointnet++}, including the across level skip links, allows PointNet++ to encode features at different scales of locality in a hierarchical manner. We follow this approach in our feature extraction network and build $\phi$ with a segmentation module consisting of a sequence of three set abstraction layers and three feature propagation layers. The output of $\phi$ are the sets of deep features $\zeta_i$, which are concatenated with the positions of the input points in $P_i$, named $\Tilde{\zeta}_i$, for further feature embedding later.

\subsubsection{Keypoint detection}
Inspired by \cite{yew20183dfeat} and \cite{lu2019deepvcp} we use a weighting layer as the keypoint detection network to learn the salience of the features extracted by $\phi$.  In this work we use the multi-layer perceptron (MLP) applied in DeepVCP, consisting of 3 fully-connected layers, followed by the \textit{softplus} activation function. The inputs to the keypoint detection layer are the sets of deep features extracted above $\zeta_i$ and the outputs are sets of scalar $w_i$ that will weight the points in the triplet loss.

\subsubsection{Deep feature embedding}
Following the design of DeepVCP once more, we employ a second deep feature embedding network to learn a more detailed descriptor of local patches for the triplet learning, which would effectively help improve the correspondence matching \cite{lu2019deepvcp}. Thus, $\chi$ is formed by one grouping layer and one PointNet layer (as described in \cite{qi2017pointnet++}) followed by the normalization in the last step. Local patches are constructed by aggregating neighbouring points using the grouping layer and used to learn a further embedding $\xi_i$ of local areas from the point-wise features $\Tilde{\zeta}_i$.

\section{EXPERIMENTS}
In this section we present the methodology followed to train the neural network architecture presented in the previous section on bathymetric data. Additionally, we assess the performance of our solution in the tasks of place recognition and correspondence matching and evaluate the results. Both the NN implementation and the dataset can be found here \footnote{\url{https://github.com/tjr16/bathy_nn_learning}}.

\subsection{The bathymetric dataset}
The data used in the experiments has been collected as part of a larger mission with a Hugin 3000 AUV equipped with a MBES Kongsberg EM 2040 and without any external navigation aid. The survey was carried out beneath the Thwaites glacier in west Antarctica and the segment used for these experiments covers approximately $10 km^2$, surveyed over $6$ hours. However, the vehicle had already navigated for approximately $5$ hours before reaching this area, which has resulted in accumulated DR drift present in the data. 

\begin{figure}[!h]
  \centering
  \includegraphics[width=\linewidth]{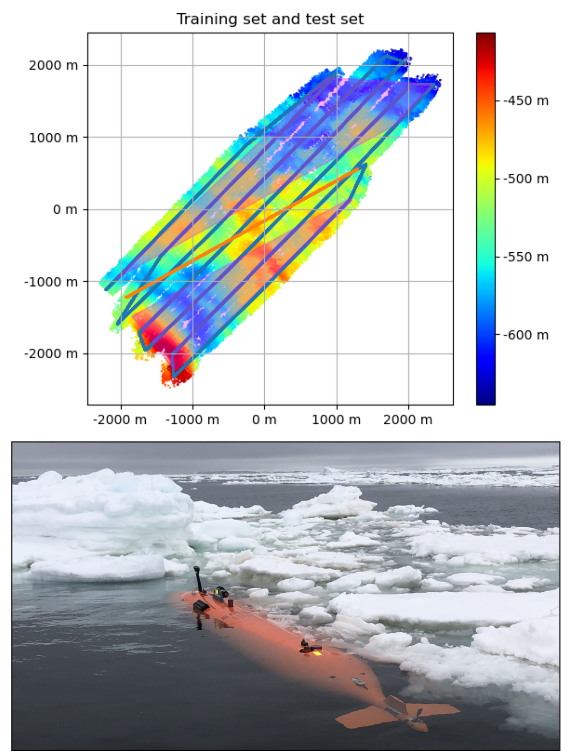}
  \caption{Bathymetry of the surveyed area used for training and testing (top) and the Hugin AUV after the survey (bottom).}
  \label{fig:bathy}
\end{figure}

\begin{figure*}[!t]
% \vspace*{-0.5cm}
  \centering
  \includegraphics[width=0.35\textwidth,angle =90]{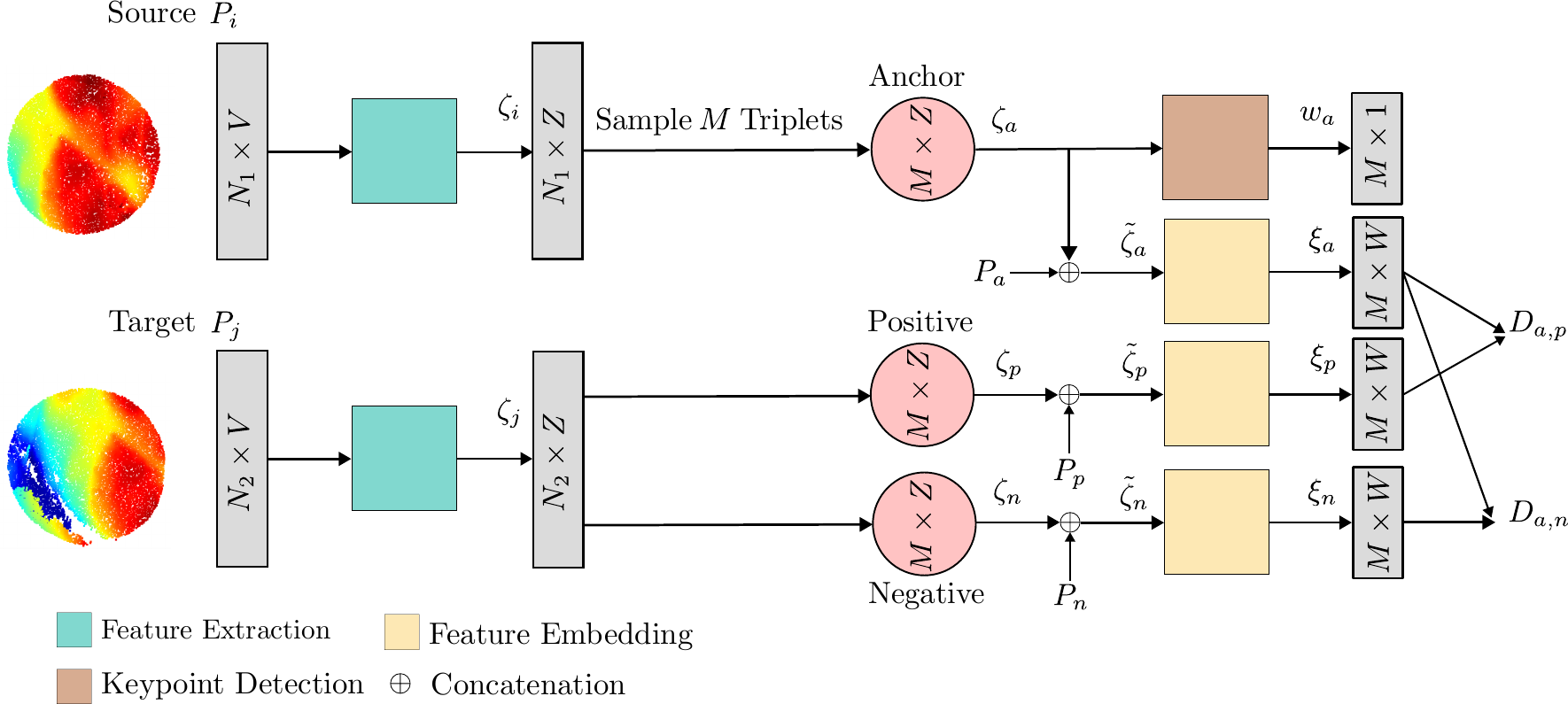}
  \caption{The Siamese architecture for training. The  feature extraction network  takes two point clouds as input and extracts informative features $\zeta$, from which we we randomly sample $M$ triplets from the overlapping area of the source and target point clouds. The features $\zeta$ from the anchor of the sampled triplets is fed into the keypoint detector to learn the saliency. In parallel, we concatenate the relative positions with $\zeta$ to get $\Tilde{\zeta}$, which is further fed into the feature embedding network to learn the final descriptor $\xi$. For a triplet, we can compute their corresponding feature descriptors $\xi_a, \xi_p, \xi_n$, which are used to construct the triplet loss. }
  \label{fig:nn-pipeline-training}
\end{figure*}
The resulting bathymetry can be seen in the top of Fig. \ref{fig:bathy}. The AUV's DR trajectory estimate has been plotted in blue for the section of the lawn mowing pattern and orange for the revisiting swath. The areas shaded in red roughly outline the sections of the data used in the training set. The swath corresponding to the orange trajectory has been used for the test set. Finally, the validation set is composed of bathymetry collected during the same mission right outside that area. The bottom image in Fig. \ref{fig:bathy} shows the Hugin after the mission. 

\subsection{Generation of triplets}
In the ideal case, metric learning for data association should be trained on point clouds collected from two crossing trajectories. However, since we only have one revisit trajectory, which we keep for the test set, we have to artificially generate overlapping point clouds for model training and validation. We do so randomly sampling cylindrical point clouds, as in \cite{dube2017segmatch}, of radius $100$ m from the red-shaded survey areas.  The resulting point clouds are grouped into positive and negative pairs based on their overlap, which can only be inferred based on the AUV's DR estimates, since ground truth is not available. A pair is considered positive if their Intersection Over Union (IoU) $\in [0.4, 0.8]$ and negative if they present no overlap. 
% add noise
We apply preprocessing and data augmentation steps to the resulting pairs in order to prepare and maximize our limited dataset. First, the point clouds are demeaned, and their absolute positions are kept as labels. We then downsample the point clouds to reduce computational requirements and obtain more uniformly distributed sets of points. Finally, to make our model more robust to errors arising from the vehicle's depth and heading, we perform data augmentation applying the following steps:\\ 
1. Each point cloud is disrupted by a random rotation $\alpha \sim \mathcal{U}(-3.0, 3.0)\degree$ along the $z$ axis.\\ % (follows a uniform distribution)
2. It is followed by a random translation across the z axis $\Delta z \sim \mathcal{U}(-0.2, 0.2)m$.\\
3. Finally, each point's coordinates are corrupted by i.i.d Gaussian noise $\delta \sim \mathcal{N}(0, 0.05)m$.

The test set, however, is created in a different manner to maximize the data available. Following the orange trajectory in Fig. \ref{fig:bathy}, we crop cylindrical point clouds every $10$ pings as target point clouds. We also sample source point clouds from nearby blue trajectories in the same way. In total, $209$ point clouds are generated following this procedure, from which $123$ positive pairs can be extracted. For comparison, we also randomly search for $123$ negative pairs among these point clouds. The size and composition of the final dataset is summarized in Table \ref{table:dataset}.

Since we seek to learn local descriptors of the geometric features within the point clouds, we do not use the created point clouds directly to construct the triplets. Instead, we crop local patches of $15$ m radius within the point clouds and look for positive samples inside the overlapping regions.
Thus, given a positive pair of point clouds, a Farthest Point Sampling (FPS) strategy is used to uniformly draw $512$ candidate anchor points from the source point cloud. From the $512$ samples, $M$ anchor points within the overlapping region are randomly selected and a local patch is cropped. This number is constrained by the hardware capacity and can be increased with more powerful resources. Finally, a patch whose distance to the anchor is below $1$ m is sampled from the target cloud as a positive and a negative one without overlap is randomly selected. To avoid cropping incomplete patches on the boundary of point clouds, the sampling is carried out in their inner area.

\begin{table}[h]
\caption{Datasets}
\label{table:dataset}
\begin{center}
\begin{tabular}{|c|c|c|c|}
\hline
Dataset & Point clouds & Positive pairs & Negative pairs\\
\hline
Training set & 1500 & 5752 & -\\
\hline
Validation set & 400 & 1614 & -\\
\hline
Test set & 209 & 123 & 123\\
\hline
\end{tabular}
\end{center}
\end{table}
% \vspace*{0.1in}
\subsection{Training of the network}
The training configuration of the architecture presented in Fig. \ref{fig:nn-pipeline-inference} is shown in Fig. \ref{fig:nn-pipeline-training}. It consists of two Siamese networks, namely, both feature extraction networks shown in Fig. \ref{fig:nn-pipeline-training} have identical weights for the source and target point clouds. The same reasoning applies to the three feature embedding networks for the anchor, positive and negative samples in the triplets. During training, instead of using every point to construct the triplets, we randomly sample $M=5$ triplets from the source point clouds every epoch and calculate the triplet loss based on these $M$ triplets. Specifically, we uniformly sample $N_1=N_2=8192$ points as input, whose $V=4$ dimensions are the combination of the 3D relative positions and the absolute depth encoding. For every point $p_k \in P$ we concatenate its relative position with $\sin{(\frac{\pi}{50} d_i)}$, where $d_i$ is the absolute depth. In this manner, the depth information will be encoded into learned feature descriptors while avoiding learning absolute depths. The dimensions of the descriptors we use are $Z=W=32$. Note that we only concatenate the learned features $\zeta$ with the relative positions of point $p$, so the dimension of $\Tilde{\zeta}$ is $32+3=35$.

To find the optimal $\theta$, we seek to optimize the loss in Eq. \ref{eq:tri_loss_ind}. However, in order to handle large amounts of triplets efficiently, we apply stochastic gradient descent (SGD) during the training. The minibatch size has been set to $\mathcal{B} = 16$ for a fair balance between computation constraints and performance. Similarly to \cite{yew20183dfeat}, we weight each individual loss $\mathcal{L}_t$ by the normalized weight across the triplets within a minibatch, $w'_{\mathcal{B}}$. Thus, the loss becomes $\mathcal{L} = \sum_{\mathcal{B}} w'_{\mathcal{B}} \mathcal{L}_{t}$ for $t \in \mathcal{B}$. We use Adam \cite{kingma2014adam} to optimize this loss, with a learning rate of $10^{-5}$ and a 5$\%$ decay per epoch.
The final model was trained in a single Nvidia GEFORCE RTX 2080 Ti GPU. It took approximately 2 hours (20 epochs) for the validation loss to converge.

\subsection{Autonomous place recognition}
We have tested our network on the task of place recognition for loop closure detection on bathymetric point clouds. The keypoint detector is used to select some interest points with high weights, and the corresponding descriptors from the feature embedding layer are used for data association. The matching between the descriptors is carried out applying a brute-force (BF) matcher with cross-check for a large number of candidates \cite{noble2016comparison}. More in detail, the similarity between features is encoded by the Euclidean distance in the latent space learned. Therefore, the BF enforces that either of the two candidate features must be the best match for the other, according to that metric, for a proposed match to be accepted. Afterwards, the correspondences with an absolute depth difference greater than 2 meters are discarded. Finally, loop closure candidates are accepted based on a minimum threshold on the number of matches.

\begin{center}
\captionof{table}{Confusion matrix with LC detections} 
\begin{tabular}{@{}cc|cc@{}}
\multicolumn{1}{c}{} &\multicolumn{1}{c}{} &\multicolumn{2}{c}{Predicted} \\ 
\multicolumn{1}{c}{} & 
\multicolumn{1}{c|}{} & 
\multicolumn{1}{c}{Positive} & 
\multicolumn{1}{c}{Negative} \\ 
\cline{2-4}
\multirow[c]{2}{*}{\rotatebox[origin=tr]{90}{Real}}
& Positive  & 27 & 96   \\[1.ex]
& Negative  & 0   & 123 \\ 
\cline{2-4}
\end{tabular}
\label{tab:confusion}
\end{center}
\vspace{0.3cm}

\begin{figure}[t]
  \centering
    \includegraphics[width=\linewidth]{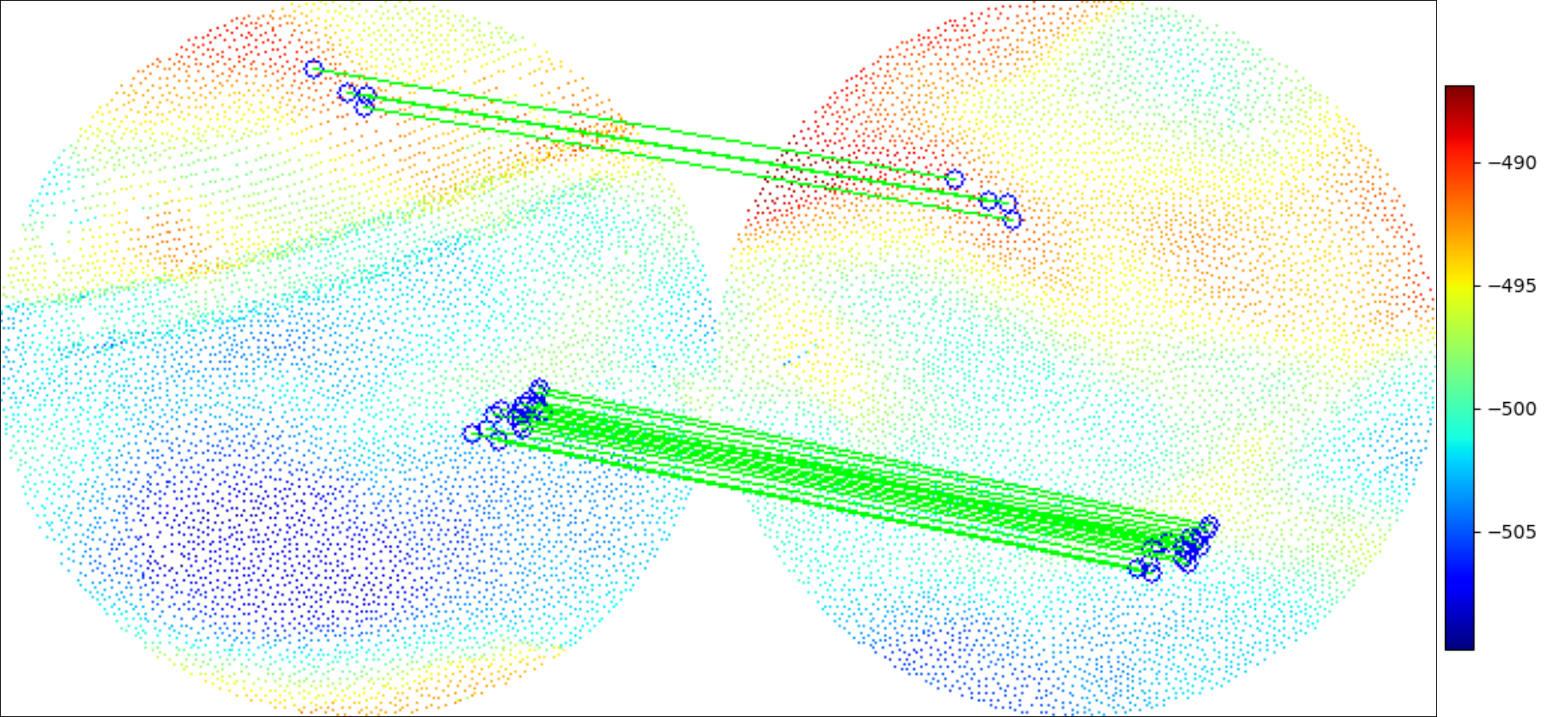}
  \caption{Instance of correspondence matching using the descriptors from our network. Depth in meters.}
  \label{fig:corresp_matching}
\end{figure}

Fig \ref{fig:corresp_matching} showcases an instance of correct correspondence matching using the descriptors from our network. For loop closure candidates to be accepted, a minimum threshold of $3$ matches has been empirically found to work well, filtering out all false positives. In the presence of more than one LC candidate for a given point cloud, the one with most matches is selected. Table \ref{tab:confusion} compiles a confusion matrix with the LC detection results for our method.
% \smallskip

\begin{figure}[h]
  \centering
  \includegraphics[width=\linewidth]{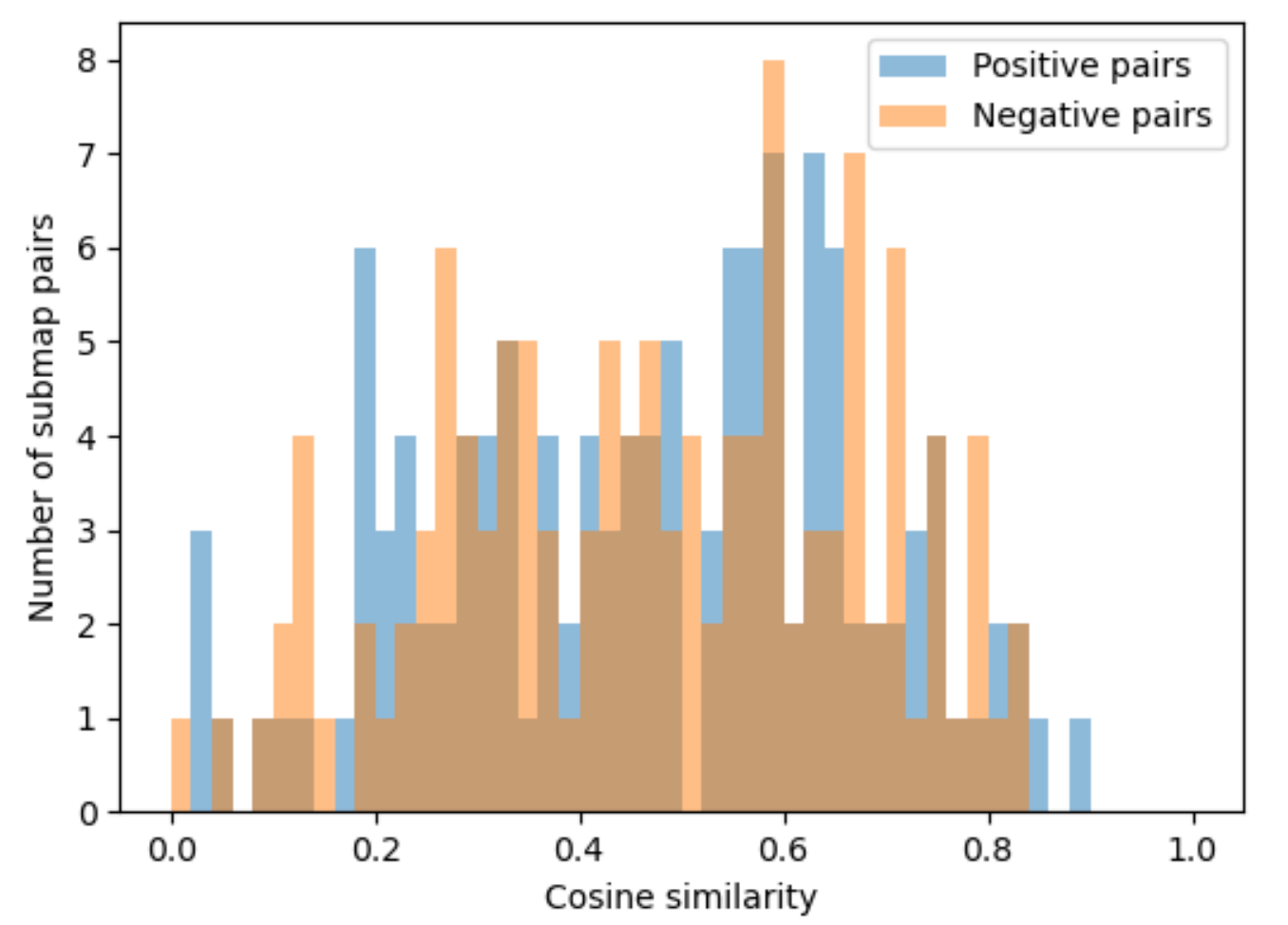}
  \caption{Cosine similarity of BoW using the approach in \cite{suresh2020active} for positive and negative pairs of submaps.}

  \label{fig:baseline_lc}
\end{figure}

To show the complexity of the task under study, we implement as a baseline method the approach presented in \cite{suresh2020active}. Their pipeline consists in a keypoint detection step based on Harris 3D, followed by the extraction of SHOT descriptors and their clustering via k-means in order to build a bag of words (BoW) per point cloud. The BoW is then used to detect loop closure candidates for point clouds registration. For a fair comparison, the k-means model has been trained on the training set beforehand.
Fig \ref{fig:baseline_lc} summarizes the results from applying the baseline to our bathymetric test set. The cosine similarity has been computed across every BoW for positive and negative matches, resulting in a mean value of $0.4712$ for positive matches and $0.4733$ for negative ones. This implies that no information has been encoded in the dictionaries and therefore they have not been further used for the LC detection experiment. This result highlights the need for specific methods for MBES point clouds from large, unstructured seabed regions.

\subsection{Coarse and fine registration of bathymetric submaps}
We now define a submap as a tuple of the form $S_i=\{P_i, T_i\}$, where $T_i\in SE(3)$ is the absolute pose of the AUV when collecting the data, rigidly attached to the set of points $P_i$. For a full view of the submap construction process from MBES pings, see \cite{torroba2019towards}.
Submap-based SLAM frameworks rely on the registration of overlapping submaps to correct the AUV poses attached to the point clouds upon revisiting an area. In this context, the correspondence matches obtained from a successful loop closure detection can be further utilized to perform a coarse registration of the target and source point clouds, thus correcting the vehicle trajectory. To this end, the singular value decomposition (SVD) of the cross-covariance matrix of the best 20 matches obtained is computed to find the rigid transformation $T^{SVD}_{i,j}$ that coarsely aligns the submaps $S_i$ and $S_j$ \cite{arun1987least}. This approach is faster than iterative methods such as least-squares solutions and is robust under reasonable levels of noise. 
Afterwards, a fine registration is carried out iteratively via GICP in order to maximize the consistency of the final point cloud reconstruction of the target area. Due to the lack of ground truth observations of the seabed areas whose bathymetry we aim to reconstruct, we choose to employ the error metric presented in \cite{roman2006consistency} to assess the performance of our network in a full registration pipeline. This metric gauges the misalignment between overlapping point clouds evaluating the vertical distances between points in the overlapping areas (which should be zero in the case of perfect alignment).
After the fine registration, the resulting, corrected relative transformation between AUV poses $T^{GICP}_{i,j}$ can be added as a constraint in a SLAM back-end \cite{torroba2020pointnetkl}.
\begin{table}[h]
\captionof{table}{Average RMS consistency errors from the DR and the submaps registration experiments} 
\begin{center}
\begin{tabular}{@{}l|llll@{}}
\cmidrule(r){1-4}
Method  & DR & Baseline & Ours &  \\ \cmidrule(r){1-4}
Bathymetric RMS (m) &  3.6405  &   3.2151      &   1.8861  &  \\ \cmidrule(r){1-4}
\end{tabular}
\end{center}
\label{tab:reg_results}
\end{table}

Table \ref{tab:reg_results} compiles the results from the registration process introduced above. It presents the average root mean square (RMS) consistency errors across the $27$ recognized positive pairs of submaps from the place recognition experiment. It can be seen how the original misalignment between overlapping point clouds, due to the drift in the AUV's DR estimate, has been reduced by approximately $50\%$ with our method, outperforming the results with the correspondences from the baseline method in \cite{suresh2020active}.

\begin{figure}[h]
  \centering
  \includegraphics[width=\linewidth]{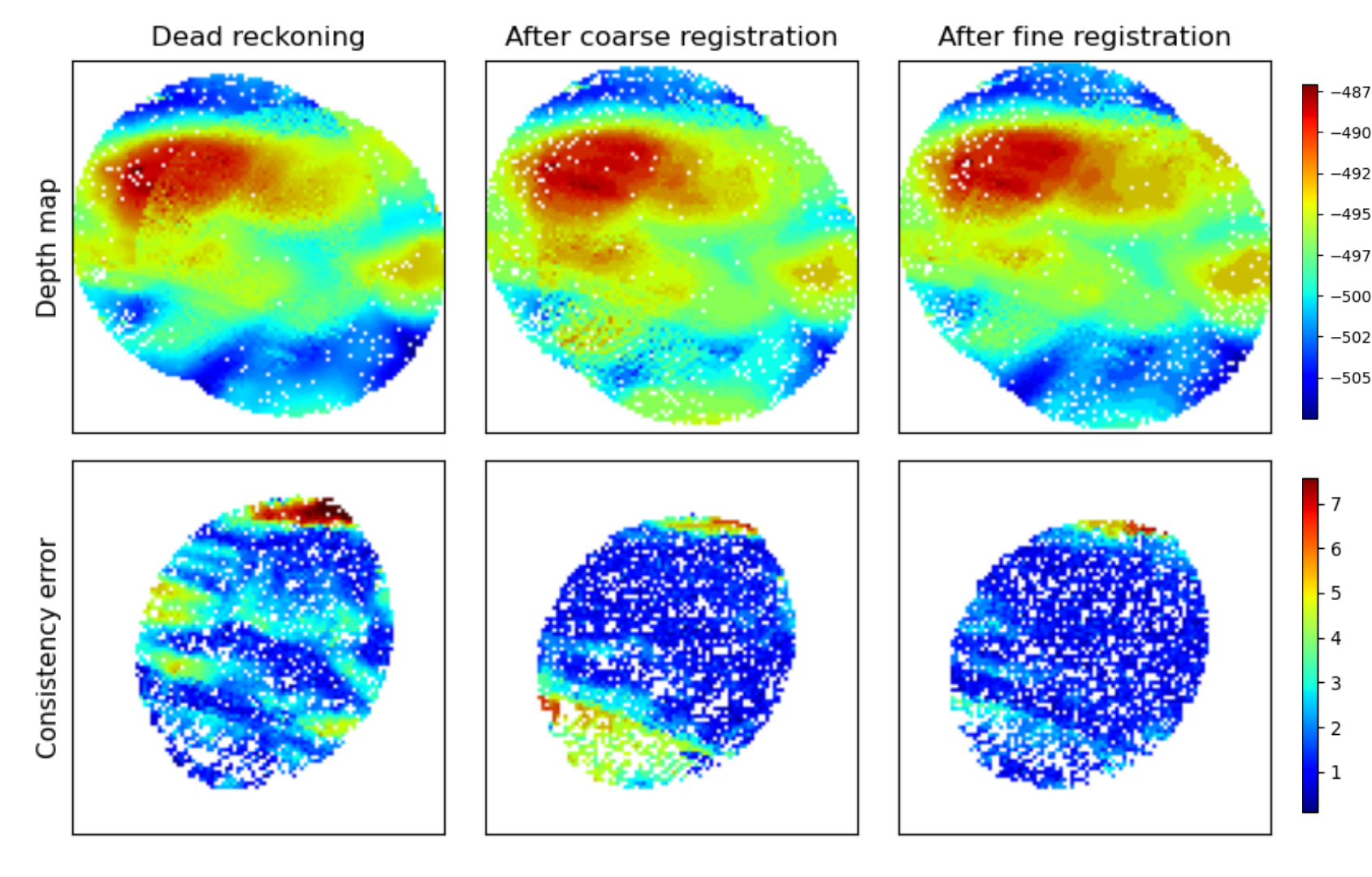}
  \caption{Depth maps (top) and consistency errors (bottom) before and after registration. Colorbars in meters.}
  \label{fig:gicp_reg}
\end{figure}

Figure \ref{fig:gicp_reg} depicts the steps in the registration process and the associated bathymetric error being minimized for a pair of submaps from the experiment. On the left, two submaps whose overlap has been detected by the network are shown with their original misalignment caused by the AUV's DR error. On the center, the submaps after a coarse registration with the features detected by the network. On the right, the final relative pose between the submaps after registration. 

% \vspace*{-1cm}

\section{CONCLUSIONS}
We have presented a learning framework for detection and description of keypoints that targets the specific challenges of bathymetric raw data for AUV SLAM. Our architecture is built upon \cite{qi2017pointnet++} and \cite{lu2019deepvcp} and applies a triplet loss \cite{schroff2015facenet} to learn the optimal metric for correspondence matching across point clouds. It has been designed to take the saliency of the points into account and to hierarchically encode local information of the geometric features within the point clouds while remaining resistant to noise. The network has been trained on bathymetric data collected with an AUV equipped with a MBES on a real under-ice mission, deprived of any external positioning system. It has been tested on the tasks of global loop closure detection and coarse submap alignment for AUV trajectory correction. Due to the lack of ground truth positioning of the vehicle, we have employed the consistency metric defined in \cite{roman2006consistency} to assess its performance. 

We show how the results presented, although limited to one dataset, outperform the baseline selected \cite{suresh2020active} on both tasks. In fact, the baseline method is quickly rendered entirely inadequate for loop closure detection in this type of terrain, which illustrates the difficulty of the problem. This reinforces our believe that specific solutions are required for data association in bathymetric data from unstructured seabed environments. Thus, to help advance research on such solutions, we make publicly available both the implementation of our architecture and the bathymetric dataset used. Additionally, the operation runtime of the network is $0.274$ seconds on average. Therefore, although not explicitly compared in the paper, this makes it a more suitable module for a real time SLAM front-end than solutions involving online learning, such as \cite{hitchcox2020point}.

% \addtolength{\textheight}{-12cm}   % This command serves to balance the column lengths
                                  % on the last page of the document manually. It shortens
                                  % the textheight of the last page by a suitable amount.
                                  % This command does not take effect until the next page
                                  % so it should come on the page before the last. Make
                                  % sure that you do not shorten the textheight too much.

%%%%%%%%%%%%%%%%%%%%%%%%%%%%%%%%%%%%%%%%%%%%%%%%%%%%%%%%%%%%%%%%%%%%%%%%%%%%%%%%

%%%%%%%%%%%%%%%%%%%%%%%%%%%%%%%%%%%%%%%%%%%%%%%%%%%%%%%%%%%%%%%%%%%%%%%%%%%%%%%%

%%%%%%%%%%%%%%%%%%%%%%%%%%%%%%%%%%%%%%%%%%%%%%%%%%%%%%%%%%%%%%%%%%%%%%%%%%%%%%%%

\section*{ACKNOWLEDGMENT}
The authors thank the Knut and Alice Wallenberg foundation for funding MUST, Mobile Underwater System Tools, project that provided the Hugin AUV for these tests.
This work was supported by Stiftelsen för Strategisk Forskning (SSF) through the Swedish Maritime Robotics Centre (SMaRC) (IRC15-0046) and by the Wallenberg AI, Autonomous
Systems and Software Program (WASP). \bibliographystyle{IEEEtran}
\bibliography{IEEEabrv,root}

\end{document}